\documentclass[12pt]{article}
\usepackage[utf8]{inputenc}
\usepackage{textcomp}
\usepackage{float}
\usepackage{graphicx}
\usepackage{amsmath}
\usepackage{amssymb}
\usepackage{hyperref}
\usepackage[margin=1in]{geometry}
\usepackage{lineno}  

\title{A Novel Spinor-Based Embedding Model for Transformers}
\author{Rick White}
\date{MiraCosta College}

\begin{document}

\maketitle

\begin{abstract}
This paper proposes a novel approach to word embeddings in Transformer models by utilizing spinors from geometric algebra. Spinors offer a rich mathematical framework capable of capturing complex relationships and transformations in high-dimensional spaces. By encoding words as spinors, we aim to enhance the expressiveness and robustness of language representations. We present the theoretical foundations of spinors, detail their integration into Transformer architectures, and discuss potential advantages and challenges.

\end{abstract}

\section{Introduction}

Word embeddings are fundamental to modern Natural Language Processing (NLP), providing a means to represent words in continuous vector spaces that capture semantic and syntactic relationships. Traditional embeddings like Word2Vec \cite{mikolov2013efficient} and GloVe \cite{pennington2014glove} have been effective but may not fully capture the complex multidimensional relationships inherent in complex data and language.\par

This paper explores the use of spinors---mathematical objects from geometric algebra---as an alternative method for word embeddings. Spinors can represent rotations and transformations in high-dimensional spaces, potentially providing richer representations for words and data. We propose integrating spinor-based embeddings into Transformer models \cite{vaswani2017attention}, which rely on self-attention mechanisms to process sequential data. This novel approach that leverages spinors from geometric algebra as an advanced form of informational embeddings.\par

Spinor embeddings may be able to provide larger semantic context between word, and also their higher-order relationships, such as contextual, temporal, hierarchical, and syntactic dependencies. Spinors offer a mathematical representation capable of handling these complex transformations in a higher-dimensional space, providing Transformer models with more expressive and nuanced language representations.\par

Unlike traditional vector embeddings, spinner embeddings operate in a multi-dimensional rotational space, capturing intricate relationships between words that go beyond mere proximity in a Euclidean space. This new embedding paradigm allows Transformer models to process data proximities more effectively, improving their ability to model dependencies across sequences, comprehend subtle patterns, and ultimately provide more accurate predictions.\par

Further these embeddings could represent a significant advancement in the field of Artificial Intelligence, offering a structured, high-dimensional approach to encoding relationships between symbolic objects. By doing so, they extend the capabilities of Transformer models, enabling them to exploit the richer, more complex geometry of linguistic data.\par

Contributions of this paper include:
\begin{itemize}
    \item Introducing spinors as a nee form of embedding model that utilizes geometric algebra for higher-dimension vector calculations.
    \item Formulating the mathematical framework for spinor embeddings.
    \item Proposing methods to integrate spinor embeddings into Transformer architectures.
    \item Discussing potential advantages and challengess.
\end{itemize}

\section{Background}

\subsection{Word Embeddings}

Word embeddings map words from a discrete vocabulary $V$ to continuous vector spaces $\mathbb{R}^n$. Formally, a word embedding is a function:

\begin{equation}
\mathbf{E}: \mathbf{V} \rightarrow \mathbb{R}^n
\end{equation}

where each word $w \in V$ is mapped to a vector $\mathbf{E}(w) \in \mathbb{R}^n$. The embedding space is trained to capture semantic and syntactic relationships, often through context-based learning objectives like predicting surrounding words.

\subsection{Geometric Algebra and Spinors}

Geometric algebra is an extension of linear algebra that provides a unified language for geometry and algebra \cite{hestenes1984clifford}. It introduces new algebraic elements called \textbf{multivectors}, which include scalars, vectors, bivectors, and higher-grade elements.

A \textbf{spinor} is an element of a Clifford algebra, a specific type of geometric algebra, which can represent rotations and transformations in space \cite{baylis2012clifford}. In $n$-dimensional space, spinors are objects that transform under spin groups, which are double covers of rotation groups $SO(n)$.

\subsubsection{Clifford Algebra}

Given a vector space $V$ over $\mathbb{R}$ with a quadratic form $Q$, the Clifford algebra $Cl(V, Q)$ \cite{clifford1882mathematical} is the associative algebra generated by $V$ subject to the relation:

\begin{equation}
v^2 = Q(v)1, \quad \forall v \in V
\end{equation}

where $1$ is the multiplicative identity.

\subsubsection{Spinors}

Spinors can be defined as elements of a minimal left ideal in the Clifford algebra. They can be represented using even-grade \textbf{multivectors} and can encode rotations via the action:

\begin{equation}
v' = RvR^{-1}
\end{equation}

where $v$ is a vector, and $R$ is a rotor (an even-grade multivector satisfying $RR^\dagger = 1$).

\section{Proposed Method}

\subsection{Spinor Representation of Words}

We propose encoding words as spinors within a suitable Clifford algebra. Each word $w \in V$ is associated with a spinor $\psi_w$ which is an element of the spinor space $S$.

\subsubsection{Embedding Function}

Define the embedding function:

\begin{equation}
\Psi: V \rightarrow S
\end{equation}

such that for each word $w$, we have:

\begin{equation}
\Psi(w) = \psi_w
\end{equation}

\subsubsection{Encoding Semantic Relationships}

Semantic relationships between words can be represented through geometric transformations. For example, analogies can be encoded using rotors:

\begin{equation}
\psi_{king} \approx R_{gender}\psi_{queen}
\end{equation}

where $R_{gender}$ is a rotor representing the gender transformation.

\subsection{Integration with Transformer Models}

Transformers utilize embeddings by mapping input tokens to continuous vectors before processing them through self-attention layers. To integrate spinor embeddings, we propose the following modifications:

\subsubsection{Embedding Layer}

Replace the traditional embedding layer with a spinor embedding layer. Each token is mapped to a spinor $\psi_w$.

\subsubsection{Spinor Operations in Self-Attention}

Adapt the self-attention mechanism to operate on spinors. Since spinors are elements of a Clifford algebra, we can define inner products and other operations necessary for attention calculations.

\subsubsection{Attention Mechanism}

The standard attention mechanism computes:

\begin{equation}
\text{Attention}(Q,K,V) = \text{softmax}\left(\frac{QK^T}{\sqrt{d_k}}\right)V
\end{equation}

where $Q, K, V$ are query, key, and value matrices.

For spinor embeddings, we define spinor inner products and use them in attention computations:

\begin{itemize}
\item Define a spinor inner product $\langle\psi_i,\psi_j\rangle$
\item Compute attention weights using spinor inner products
\end{itemize}

\subsubsection{Positional Encoding}

Since Transformers rely on positional encoding to capture sequence order, we need to define spinor-based positional encodings. We can represent positions as rotors:

\begin{equation}
\psi_w^{(p)} = R_p\psi_w
\end{equation}

where $R_p$ is a rotor corresponding to position $p$.

Note that due to the geometric nature of spinors, the spinor embeddings geometric distances are calculated using the geometric and the inner products, not the dot products for calculating similarities.

\subsection{Spinor Embedding Analysis}

Spinor embeddings generally have higher dimensional complexity than standard word embeddings. Here's why:

\begin{itemize}
\item \textbf{Representation Space}: While a standard word embedding in $\mathbb{R}^n$ has $n$ dimensions, a spinor embedding in $Cl(p, q)$ can represent up to $2^n$ different components (scalar, vector, bivector, etc., up to the pseudoscalar). This exponential relationship means that spinor embeddings can potentially represent much more information.

\item \textbf{Degrees of Freedom}: Spinor embeddings can capture more complex relationships and transformations within the same base dimensionality. A 4D spinor space, for example, can represent rotations in 3D space, something that would require more dimensions in a standard vector embedding.

\item \textbf{Computational Complexity}: Operations in spinor space, particularly the geometric product, can be more computationally intensive than simple vector operations. The geometric product has a complexity of $O(2^n)$ in the general case, although this can be optimized for specific algebras.

\item \textbf{Memory Requirements}: In the worst case, a spinor embedding might require $2^n$ parameters per word, compared to $n$ parameters for standard embeddings. However, in practice, many of these components may be zero or have specific structures that allow for more efficient representations.

\item \textbf{Expressive Power}: The higher dimensional complexity of spinor embeddings allows them to capture more nuanced relationships and transformations. This increased expressiveness is one of the main motivations for using spinor embeddings.
\end{itemize}

However, it's important to note that this higher dimensional complexity doesn't necessarily mean that spinor embeddings always use more memory or computation in practice:

\begin{itemize}
\item \textbf{Sparsity}: Many applications of spinor embeddings might use sparse representations, where only a few of the $2^n$ possible components are non-zero.

\item \textbf{Structured Representations}: The algebraic structure of spinors often allows for more compact representations of certain transformations.

\item \textbf{Efficiency-Expressiveness Trade-off}: The higher dimensional complexity of spinors might allow for more efficient representation of certain linguistic phenomena, potentially requiring fewer base dimensions than standard embeddings to achieve similar expressiveness.
\end{itemize}

While spinor embeddings do have higher dimensional complexity in theory, their practical implementation can be optimized to mitigate some of the computational and memory costs. The higher complexity is a trade-off for increased expressiveness and the ability to represent more complex relationships in a geometrically meaningful way. Whether this trade-off is beneficial depends on the specific requirements of the NLP task at hand and the computational resources available.

\subsection{Examples of Geometric Interpretations of Spinor Embeddings}

Here are some examples of how rotations and reflections in spinor space can be interpreted in the context of natural language processing.

\subsubsection{Example \#1: Rotation - Actor to Actress}

\begin{figure}[h]
\centering
\includegraphics[width=0.6\textwidth]{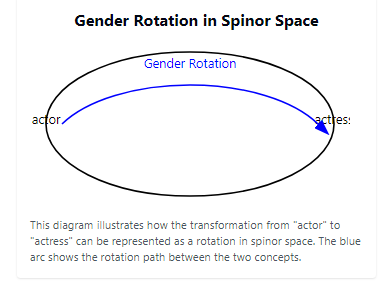}
\caption{Rotation from "actor" to "actress" in spinor space}
\label{fig:actor_actress_rotation}
\end{figure}

In this example, a rotation in spinor space smoothly transforms "actor" to "actress". This same rotation could potentially be applied to other word pairs like "waiter" → "waitress" or "prince" → "princess".

\textbf{360-Degree vs 720-Degree Rotation}: In our original example, we showed a simple 180-degree rotation from "actor" to "actress". However, if we continue this rotation, we find that:

\begin{itemize}
\item A 360-degree rotation doesn't return to "actor", but to a state we might call "actor*"
\item It takes a full 720-degree rotation to truly return to the original "actor" state
\end{itemize}

\textbf{Four States}: In a complete 720-degree rotation, we actually pass through four states: actor → actress → actor* → actress* → actor
Where the * states are mathematically distinct from the non-* states, even though they might appear identical in our normal 3D space.

\textbf{Phase Information}: The difference between "actor" and "actor*" (or "actress" and "actress*") is in their phase. This phase difference isn't observable in regular space but is a crucial part of the spinor representation.

\textbf{Quantum Mechanical Analogy}: This behavior is analogous to the quantum mechanical property of fermions (like electrons), where a 360-degree rotation changes the sign of the wavefunction, and it takes a 720-degree rotation to return to the original state.

\begin{figure}[h]
\centering
\includegraphics[width=0.6\textwidth]{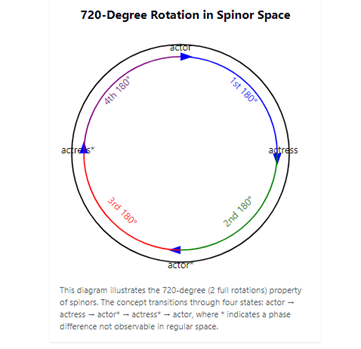}
\caption{720-degree rotation in spinor space}
\label{fig:720_degree_rotation}
\end{figure}

\textbf{Implications for NLP}: In the context of natural language processing, this property could potentially be used to represent more nuanced relationships between words or concepts. For example:

\begin{itemize}
\item The transition from "actor" to "actress" might represent a change in grammatical gender.
\item The transition from "actor" to "actor*" could represent a more subtle change, perhaps in connotation or context, that isn't captured by traditional vector embeddings.
\end{itemize}

This 720-degree property of spinors allows for richer, more complex representations of linguistic concepts and relationships. It provides a way to encode information that goes beyond what's possible with traditional vector embeddings, potentially capturing subtle nuances in meaning or usage that are difficult to represent otherwise.

\subsubsection{Example \#2: Rotation - Verb Tenses}

Rotations can also represent changes in verb tenses:

\begin{figure}[H]
\centering
\includegraphics[width=0.5\textwidth]{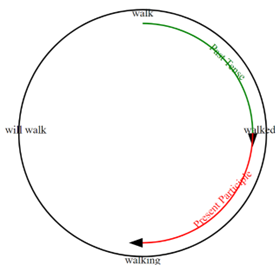}
\caption{Rotation representing verb tense changes}
\label{fig:verb_tense_rotation}
\end{figure}

Here, rotations in spinor space represent transitions between different verb tenses. A 90° rotation might transform "walk" to "walked" (past tense), while a 180° rotation could represent the change to "walking" (present participle).

\subsubsection{Example \#3: Reflections - Antonyms}

Reflections can be used to represent relationships between antonyms: 

\begin{figure}[H]
\centering
\includegraphics[width=0.6\textwidth]{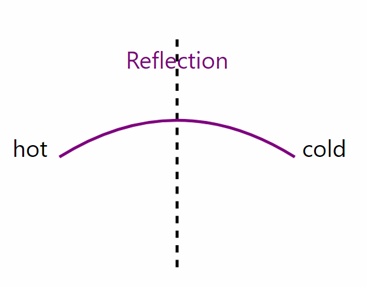}
\caption{Reflection representing antonyms}
\label{fig:antonym_reflection}
\end{figure}

In this example, a reflection transforms "hot" to "cold". This operation could potentially be applied to other antonym pairs like "big" $\leftrightarrow$ "small" or "fast" $\leftrightarrow$ "slow".

\subsubsection{Example \#4: Reflections - Negation}

Reflections can also represent negation in language: 

\begin{figure}[H]
\centering
\includegraphics[width=0.7\textwidth]{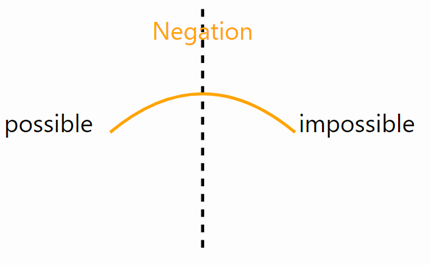}
\caption{Reflection representing negation}
\label{fig:negation_reflection}
\end{figure}

A reflection operation represents the transformation from "possible" to "impossible". This same operation could potentially be applied to transform "logical" to "illogical" or "relevant" to "irrelevant".

\subsubsection{Example \#5: Compound Transformation (Reflections and Rotation) - Combining Tense and Negation}

One of the powerful aspects of spinor embeddings is the ability to compose transformations. We can combine a rotation (tense change) with a reflection (negation):

\begin{figure}[H]
\centering
\includegraphics[width=0.6\textwidth]{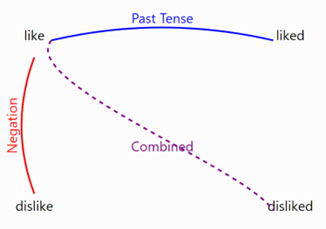}
\caption{Compound transformation combining tense change and negation}
\label{fig:compound_transformation}
\end{figure}

In this example, we see how a rotation (changing "like" to "liked") can be combined with a reflection (changing "like" to "dislike") to produce a compound transformation ("like" to "disliked").

These geometric interpretations demonstrate how spinor embeddings can represent complex linguistic relationships and transformations in an intuitive, geometrically meaningful way. The ability to compose these transformations (through multiplication of spinors) allows for representing complex linguistic phenomena in a mathematically elegant manner.

\subsection{Geometric Operations}

These examples illustrate how geometric operations in spinor space can be interpreted in the context of natural language processing:

\begin{itemize}
\item \textbf{Rotations} can represent smooth transitions between related concepts or grammatical transformations. For instance:
    \begin{itemize}
    \item Gender changes in nouns (e.g., "actor" to "actress")
    \item Verb tense changes (e.g., "walk" to "walked" to "walking")
    \end{itemize}
\item \textbf{Reflections} can represent more abrupt changes or opposites, such as:
    \begin{itemize}
    \item Antonyms (e.g., "hot" to "cold")
    \item Negations (e.g., "possible" to "impossible")
    \end{itemize}
\item \textbf{Compound Transformations} demonstrate the power of composing multiple operations, like combining a tense change with negation (e.g., "like" to "disliked").
\end{itemize}

The key advantages of these geometric interpretations include:

\begin{itemize}
\item \textbf{Intuitive Representation}: Complex linguistic relationships can be visualized and understood geometrically.
\item \textbf{Compositionality}: Multiple transformations can be easily combined through spinor multiplication.
\item \textbf{Consistency}: The same transformation can often be applied across multiple word pairs, capturing linguistic patterns.
\item \textbf{Continuous Space}: Rotations allow for representing gradual changes or degrees of transformation.
\end{itemize}

These geometric interpretations make spinor embeddings particularly powerful for tasks involving analogical reasoning, grammar transformation, or capturing nuanced relationships between words. However, it's important to note that while these geometric interpretations are theoretically powerful, realizing their full potential in practical NLP applications will require active research.

\section{Theoretical Advantages}

\subsection{Expressiveness}

\subsubsection{Spinor Expressiveness in Natural Language Processing}

Spinor expressiveness refers to the rich representational capacity of spinors compared to traditional vector embeddings. This increased expressiveness allows spinors to capture more complex and nuanced relationships in language.

\subsubsection{Key Aspects of Spinor Expressiveness:}

\begin{itemize}
\item \textbf{Higher-Dimensional Information:} While a vector in n-dimensional space has n components, a spinor in the same space can represent $2^n$ degrees of freedom. This allows for encoding much more information within the same base dimensionality.

\item \textbf{Complex Transformations:} Spinors can naturally represent rotations and other complex transformations in high-dimensional spaces. This is particularly useful for modeling the multifaceted nature of language.

\item \textbf{Phase Information:} Spinors carry phase information, which doesn't have a direct analogue in traditional vector representations. This phase can potentially encode subtle aspects of meaning or context.

\item \textbf{Quantum-Inspired Superposition:} Spinors can represent multiple states simultaneously, similar to quantum superposition. This property can be useful for modeling ambiguity or multiple meanings in language.

\item \textbf{Continuous and Discrete Representations:} Spinors can smoothly transition between discrete states, allowing for representation of both categorical and continuous linguistic phenomena.
\end{itemize}

\subsubsection{Comparison with Traditional Vector Embeddings:}

\begin{itemize}
\item \textbf{Rotation vs. Translation:} Vector models often represent relationships as translations (e.g., king - man + woman = queen). Spinor models can represent these as rotations, which can be more natural for certain types of relationships.

\item \textbf{Non-Linear Transformations:} While vector models typically use linear transformations, spinor models can easily incorporate non-linear transformations through rotations and other operations.

\item \textbf{Entanglement:} Spinors can represent entangled states, allowing for modeling of complex dependencies between different aspects of language that are difficult to capture with independent vector components.
\end{itemize}

\subsubsection{Visual Representation:}

\begin{figure}[h]
\centering
\includegraphics[width=0.8\textwidth]{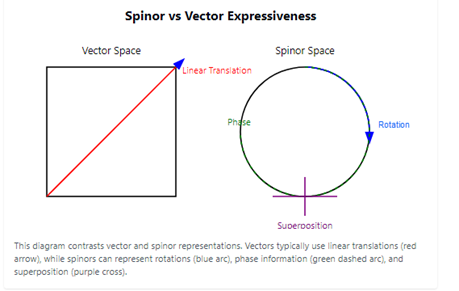}
\caption{Comparison of vector and spinor embedding expressiveness}
\label{fig:expressiveness_comparison}
\end{figure}

In summary, spinors can represent complex transformations and relationships, potentially capturing nuances in language that traditional vectors cannot.

\subsection{Rotational Invariance}

Spinors naturally encode rotational symmetries. In language, certain transformations (e.g., syntactic variations) could be modeled as rotations in spinor space.

\subsubsection{Advantages of Rotational Invariance in Spinor Embeddings}

\paragraph{Preservation of Semantic Relationships}

Rotational invariance ensures that the relative relationships between words remain consistent regardless of the specific orientation in the embedding space. This property is particularly valuable for maintaining semantic relationships:

\begin{itemize}
\item \textbf{Analogy Preservation}: Relationships like "king" is to "queen" as "man" is to "woman" are preserved under rotations, making these relationships more robust and consistent across the embedding space.

\item \textbf{Semantic Clustering}: Words with similar meanings tend to cluster together, and rotational invariance ensures these clusters maintain their relative positions regardless of the embedding space's orientation.
\end{itemize}

\paragraph{Improved Generalization}

Rotational invariance can lead to better generalization of the model:

\begin{itemize}
\item \textbf{Reduced Overfitting}: The model is less likely to learn spurious correlations based on arbitrary orientations in the embedding space, potentially leading to better generalization to unseen data.

\item \textbf{Data Augmentation}: Rotational invariance implicitly provides a form of data augmentation, as the model effectively sees multiple orientations of the same semantic relationships during training.
\end{itemize}

\paragraph{Stability Across Languages and Domains}

When dealing with multilingual or cross-domain tasks, rotational invariance can provide additional benefits:

\begin{itemize}
\item \textbf{Cross-lingual Embeddings}: Rotational invariance can help in aligning embeddings across different languages, as the relative positions of concepts should remain consistent regardless of the specific embedding space orientation.

\item \textbf{Domain Transfer}: When transferring models between domains, rotational invariance can help maintain the learned semantic relationships, even if the overall orientation of the embedding space changes.
\end{itemize}

\paragraph{Geometric Interpretability}

Spinor embeddings with rotational invariance offer a more geometrically interpretable representation:

\begin{itemize}
\item \textbf{Meaningful Transformations}: Rotations in the spinor space can represent meaningful transformations (e.g., changing tense, plurality, or other grammatical features) in a way that is consistent across the entire embedding space.

\item \textbf{Hierarchical Relationships}: The geometric nature of spinors allows for representing hierarchical relationships in a more natural way, potentially capturing complex linguistic structures.
\end{itemize}

\paragraph{Robustness to Initialization and Training Variations}

Rotational invariance can provide robustness to various factors in model training:

\begin{itemize}
\item \textbf{Initialization Invariance}: The model's performance becomes less dependent on the initial random orientation of the embedding space, potentially leading to more consistent results across different training runs.

\item \textbf{Optimization Stability}: Rotational invariance can help stabilize the optimization process, as the loss landscape becomes more consistent under rotations.
\end{itemize}

\paragraph{Efficient Representation of Symmetries}

Spinors are particularly efficient at representing symmetries and transformations:

\begin{itemize}
\item \textbf{Compact Representation}: Complex transformations that might require multiple operations in traditional vector spaces can often be represented more compactly using spinors.

\item \textbf{Continuity of Transformations}: Spinors can represent continuous transformations smoothly, which can be beneficial for modeling gradual changes in meaning or grammatical features.
\end{itemize}

\paragraph{Enhanced Compositionality}

The rotational properties of spinors align well with the compositional nature of language:

\begin{itemize}
\item \textbf{Phrase and Sentence Embeddings}: Combining word embeddings to form phrase or sentence embeddings can potentially be more semantically consistent when using rotationally invariant spinor representations.

\item \textbf{Recursive Structures}: The ability to compose transformations naturally in spinor space could be particularly useful for representing recursive linguistic structures.
\end{itemize}

\subsection{Compositionality}

\subsubsection{Spinor Compositionality in Natural Language Processing}

Compositionality in the context of spinor embeddings refers to the ability to combine multiple linguistic transformations by multiplying their corresponding spinors. This property allows for the representation of complex language phenomena as a series of simple geometric operations.

\subsubsection{Key Aspects of Spinor Compositionality:}

\begin{itemize}
\item \textbf{Multiplicative Nature}: Spinor transformations are combined through multiplication, not addition. This is in contrast to traditional vector embeddings, where transformations are often represented as vector addition.

\item \textbf{Non-Commutativity}: The order of multiplication matters. A * B is not necessarily equal to B * A, which allows for capturing the importance of sequence in language transformations.

\item \textbf{Preservation of Structure}: Spinor multiplication preserves the algebraic structure of the spinor space, ensuring that the result of combining transformations is still a valid spinor.

\item \textbf{Efficiency}: Complex transformations can often be represented more compactly using spinor multiplication than with traditional vector operations.
\end{itemize}

\subsubsection{Example in Language Processing:}

Let's consider how we might represent the transformation from "walk" to "walked quickly" using spinor compositionality:

\begin{itemize}
\item Let $S_{past}$ be the spinor representing the transformation to past tense.
\item Let $S_{adverb}$ be the spinor representing the addition of the adverb "quickly".
\end{itemize}

The complete transformation can be represented as:

\begin{equation}
\text{"walked quickly"} = S_{adverb} * S_{past} * \text{"walk"}
\end{equation}

This composition captures both the change in tense and the addition of the adverb in a single operation.

\subsubsection{Visual Representation:}

\begin{figure}[h]
\centering
\includegraphics[width=0.6\textwidth]{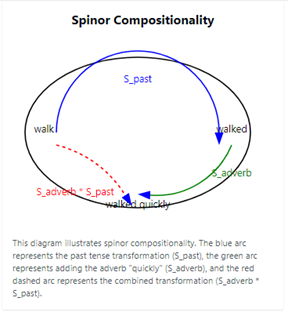}
\caption{Spinor compositionality in language transformations}
\label{fig:spinor_compositionality}
\end{figure}

Spinors can be combined through multiplication, more efficiently aligning with the compositional nature of language:

\begin{equation}
\psi_{\text{phrase}} = \psi_{w_1}\psi_{w_2}\cdots\psi_{w_n}
\end{equation}

\section{Mathematical Framework}

\subsection{Constructing the Spinor Space}

Choose an appropriate Clifford algebra $Cl(p, q)$, where $p + q = n$, the dimensionality of the space.

\subsubsection{Basis Vectors}

Let $\{e_1,e_2,\ldots,e_n\}$ be an orthonormal basis for $\mathbb{R}^{p,q}$ with the metric:

\begin{equation}
e_i^2 = \begin{cases}
+1, & 1 \leq i \leq p \\
-1, & p+1 \leq i \leq n
\end{cases}
\end{equation}

\subsubsection{Multivectors}

\begin{equation}
M = \sum_{i_1 < \cdots < i_k}^n a_{i_1 \cdots i_k} e_{i_1} \wedge \cdots \wedge e_{i_k}
\end{equation}

An element of the Clifford algebra is a multivector:

\begin{equation}
A = \sum_{k=0}^n \sum_{i_1<\cdots<i_k} a_{i_1\cdots i_k} e_{i_1}\cdots e_{i_k}
\end{equation}

where $a_{i_1\cdots i_k} \in \mathbb{R}$.

\subsection{Spinor Embeddings}

Each word is mapped to a spinor constructed from the Clifford algebra.

\subsubsection{Generating Spinors}

One method to generate spinors for words is to use exponentials of bivectors:

\begin{equation}
\psi_w = \exp(B_w)
\end{equation}

where $B_w$ is a bivector associated with word $w$.

\subsubsection{Rotor Representation}

Rotors are spinors of the form:

\begin{equation}
R = \exp\left(-\frac{\theta}{2}B\right)
\end{equation}

where $\theta$ is the angle of rotation, and $B$ is a unit bivector.

\subsection{Inner Product in Spinor Space}

Define an inner product on spinors to compute attention weights.

\subsubsection{Dirac Inner Product}

For spinors $\psi$ and $\phi$, the Dirac inner product is:

\begin{equation}
\langle\psi,\phi\rangle = \psi^\dagger\phi
\end{equation}

where $\psi^\dagger$ is the reversion (complex conjugate transpose) of $\psi$.

\subsection{Attention Computation with Spinors}

Modify the attention mechanism to use spinor inner products:

\begin{equation}
\text{Attention}(\psi_q,\psi_k,\psi_v) = \text{softmax}\left(\frac{\langle\psi_q,\psi_k\rangle}{\sqrt{d_s}}\right)\psi_v
\end{equation}

where $\psi_q,\psi_k,\psi_v$ are the query, key, and value spinors, and $d_s$ is the dimensionality of the spinor space.

\section{Experiments}

\subsection{Baseline Comparison}

Train Transformer models on standard NLP tasks (e.g., language modeling, translation) using both traditional embeddings and spinor embeddings.

\subsubsection{Dataset}

Use datasets like WMT \cite{wmt} for translation tasks or WikiText \cite{wikitext} for language modeling.

\subsubsection{Evaluation Metrics}

Evaluate using BLEU scores \cite{papineni2002bleu} for translation and perplexity for language modeling.

\subsection{Ablation Studies}

Investigate the effect of spinor dimensionality and algebraic properties on performance.

\subsubsection{Varying Clifford Algebra}

Test different Clifford algebras $Cl(p,q)$ to find the optimal configuration.

\subsubsection{Dimensionality Reduction}

Experiment with techniques like PCA \cite{pearson1901liii} to reduce the dimensionality of spinor embeddings.

\subsection{Visualization}

Visualize word relationships in spinor space using projections or by interpreting geometric transformations.

\subsubsection{Projection Techniques}

Use techniques like t-SNE \cite{maaten2008visualizing} or UMAP \cite{mcinnes2018umap} adapted for multivector spaces.

\section{Challenges and Considerations}

\subsection{Computational Complexity}

Spinor operations may be computationally intensive. Efficient algorithms and approximations may be necessary.

\subsubsection{Algorithm Optimization}

Implement efficient geometric algebra libraries or leverage GPU acceleration for computations. Techniques from \cite{dorst2007geometric} can be applied to optimize spinor operations.

\subsection{Dimensionality}

Choosing the appropriate dimensionality and Clifford algebra is critical. Higher dimensions may capture more nuances but increase computational load.

\subsubsection{Trade-Off Analysis}

Balance between expressiveness and computational feasibility. Consider using dimensionality reduction techniques like PCA \cite{pearson1901liii} to find an optimal representation.

\subsection{Interpretability}

Spinors are less intuitive than vectors. Developing intuitive interpretations of spinor embeddings is a challenge.

\subsubsection{Educational Tools}

Develop visualization tools to aid understanding of spinor operations. Techniques like t-SNE \cite{maaten2008visualizing} or UMAP \cite{mcinnes2018umap} could be adapted for spinor spaces to help visualize word relationships.

\section{Conclusion}

Integrating spinors into word embeddings offers a novel approach with the potential to enhance the capabilities of Transformer models. While theoretical advantages are promising, empirical validation is necessary. Future work will focus on implementing spinor embeddings in practical models and evaluating their performance across various NLP tasks.

\section{Future Work}

\begin{itemize}
\item \textbf{Empirical Evaluation:} Implement and test spinor embeddings in real-world NLP tasks, comparing their performance against traditional vector embeddings across a range of benchmark datasets.
\item \textbf{Optimization Strategies:} Develop methods to reduce computational overhead, potentially through sparse representations or approximation techniques.
\item \textbf{Theoretical Analysis:} Further explore the mathematical properties of spinor embeddings in language modeling, including their ability to capture complex linguistic phenomena.
\item \textbf{Cross-lingual Applications:} Investigate the potential of spinor embeddings in multilingual models and cross-lingual transfer learning tasks.
\item \textbf{Integration with Other AI Models:} Explore the application of spinor embeddings in other areas of AI beyond NLP, such as computer vision or multimodal learning.
\end{itemize}

\appendix

\section{Spinor Operations}

\subsection{Multiplication}

Spinor multiplication is associative but not necessarily commutative. For spinors $\psi$ and $\phi$:

\begin{equation}
\psi\phi \neq \phi\psi
\end{equation}

\subsection{Reversion}

The reversion of a multivector $M$ is defined by reversing the order of vectors in each component:

\begin{equation}
M = \sum_{k=1}^n (-1)^{k(k-1)/2} \sum_{i_1<\cdots<i_k} a_{i_1\cdots i_k} e_{i_1} \wedge \cdots \wedge e_{i_k}
\end{equation}

\subsection{Norm}

The norm of a spinor $\psi$ can be defined using the scalar part of $\psi^\dagger\psi$:

\begin{equation}
\|\psi\|^2 = \langle\psi^\dagger\psi\rangle_0
\end{equation}

where $\langle\cdot\rangle_0$ denotes the scalar (grade-0) part.

\subsection{Exponential Map}

The exponential of a bivector $\mathbf{B}$ is given by:

\begin{equation}
\exp(\mathbf{B}) = \cosh(\|\mathbf{B}\|) + \frac{\mathbf{B}}{\|\mathbf{B}\|}\sinh(\|\mathbf{B}\|)
\end{equation}

for hyperbolic rotations (when $\mathbf{B}^2 > 0$), or:

\begin{equation}
\exp(\mathbf{B}) = \cos(\|\mathbf{B}\|) + \frac{\mathbf{B}}{\|\mathbf{B}\|}\sin(\|\mathbf{B}\|)
\end{equation}

for circular rotations (when $\mathbf{B}^2 < 0$).

\section{Implementation Details}

\subsection{Computational Techniques}

To handle the computational complexity, we can utilize:

\begin{itemize}
\item \textbf{Sparse Representations:} Represent multivectors sparsely to reduce memory usage.
\item \textbf{Efficient Algorithms:} Use algorithms optimized for geometric algebra operations \cite{dorst2007geometric}.
\item \textbf{Approximation Methods:} Employ techniques like low-rank approximations or randomized algorithms to speed up computations.
\end{itemize}

\subsection{Integration with Deep Learning Frameworks}

Implement spinor embeddings within existing deep learning frameworks (e.g., PyTorch, TensorFlow) by defining custom layers and operations.

\subsubsection{Custom Layers}

Create layers that handle spinor arithmetic and can be integrated into the model's computational graph.

\subsubsection{Autograd Compatibility}

Ensure that operations are compatible with automatic differentiation for backpropagation.

\section{Summary}

The introduction of spinors into word embeddings opens new avenues for representing linguistic information. By leveraging the rich structure of geometric algebra, we can potentially capture deeper semantic and syntactic relationships. While challenges exist, particularly in computational efficiency and interpretability, the proposed method offers a promising direction for future research in NLP and deep learning.

This approach could lead to more expressive and geometrically meaningful representations of language, potentially improving performance on a wide range of NLP tasks. Future work will focus on empirical validation, optimization, and exploring applications beyond traditional NLP domains.

\end{document}